\documentclass[10pt,twocolumn,letterpaper]{article}

 \usepackage[pagenumbers]{cvpr} 

\usepackage[dvipsnames]{xcolor}

\definecolor{cvprblue}{rgb}{0.21,0.49,0.74}
\usepackage[pagebackref,breaklinks,colorlinks,citecolor=cvprblue]{hyperref}

\def\paperID{12851} 
\def\confName{CVPR}
\def\confYear{2024}

\title{Normalizing Flows on the Product Space of SO(3) Manifolds\\for Probabilistic Human Pose Modeling}
\author{
Olaf Dünkel$^1{}^2$, Tim Salzmann$^3$, Florian Pfaff $^4$ \\
$^1$ Max Planck Institute for Informatics, \quad $^2$ Karlsruhe Institute of Technology, \\
$^3$ Technical University of Munich,\quad $^4$ University of Stuttgart \\
{\tt\small oduenkel@mpi-inf.mpg.de}, {\tt\small tim.salzmann@tum.de}, {\tt\small pfaff@ias.uni-stuttgart.de}
}

\begin{document}
\maketitle
\begin{abstract}

Normalizing flows have proven their efficacy for density estimation in Euclidean space, but their application to rotational representations, crucial in various domains such as robotics or human pose modeling, remains underexplored. 
Probabilistic models of the human pose can benefit from approaches that rigorously consider the rotational nature of human joints.
For this purpose, we introduce \textit{HuProSO3}, a normalizing flow model that operates on a high-dimensional product space of SO(3) manifolds, modeling the joint distribution for human joints with three degrees of freedom.
HuProSO3's advantage over state-of-the-art approaches is demonstrated through its superior modeling accuracy in three different applications and its capability to evaluate the exact likelihood.
This work not only addresses the technical challenge of learning densities on SO(3) manifolds, but it also has broader implications for domains where the probabilistic regression of correlated 3D rotations is of importance. Code will be available at \url{https://github.com/odunkel/HuProSO}.

\end{abstract}

\section{Introduction}
\label{sec:intro}
Modeling uncertainties in high-dimensional Euclidean product spaces is a well-explored research problem \cite{DBLP:journals/corr/DinhKB14,DBLP:conf/iclr/DinhSB17,NEURIPS2018_d139db6a,NEURIPS2019_7ac71d43,Papamakarios2019NormalizingFF,9089305}. However, these methods often fall short in representing problems with an inherent rotational nature. This limitation is evident in scenarios like correlated object rotations in a scene or the orientation of animals in swarm behaviors, where the problems are modeled by a product space of SO(3) manifolds.
A particularly relevant example is human pose modeling, a problem of multiple correlated joint rotations.  
Accurately modeling human poses as densities on the Cartesian product of such joint rotations holds significant value for various fields, including computer vision and robotics.

\begin{figure}[t] 
	\centering
	\includegraphics[width=\columnwidth]{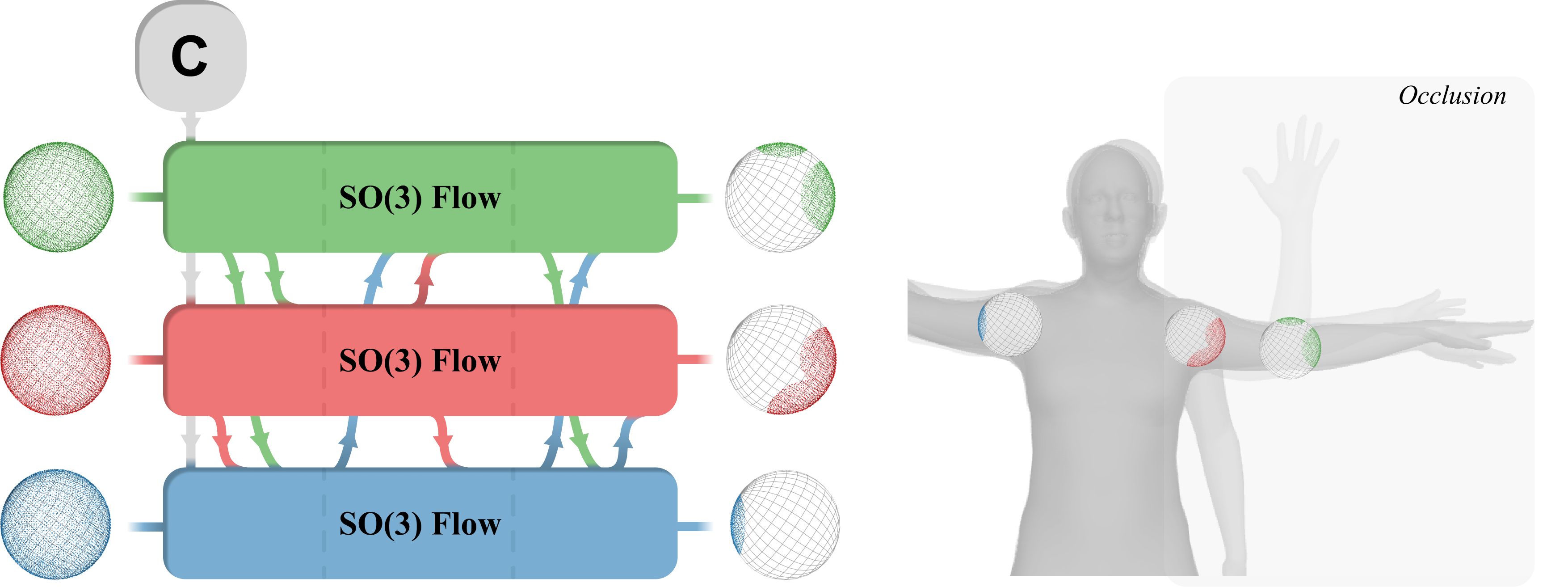}

	\caption{Application of the normalizing flow for occcluded joints. Left: A normalizing flow defined on SO(3) manifolds enables the learning of expressive human pose distributions, incorporating a context vector $\mathbf{c}$ for conditioning. Right: Renderings of probable poses given condition $\mathbf{c}$, which is the observation of the human with the left arm occluded. The right arm's pose is estimated with high certainty, while the left arm demonstrates varied but realistic poses due to the occlusion.}
	\label{fig:hero}
    \vspace{-10pt}
\end{figure}

A probabilistic model defined on SO(3) manifolds representing joint rotation distributions is relevant for human pose and motion estimation. This model acts as a reliable prior in scenarios with unclear, noisy, incomplete, or absent observations \cite{LEE1985148,kolotouros_probabilistic_2021,sengupta_humaniflow_2023,ci_gfpose_2022,salzmann_motron_2022}, with an example being depicted in \cref{fig:hero}. 
Additionally, such a model is crucial in human-robot interaction for implementing risk-aware planning and control, necessitating the representation of human movements as normalized probabilistic densities \cite{DBLP:journals/corr/abs-2009-05702}.

Normalizing flows are recognized for their capability to learn normalized densities. Yet, their application has predominantly been in learning joint densities within Euclidean spaces, not fully addressing the properties of human pose, which is significantly influenced by the rotational nature of the human joints. 
Although some previous approaches in probabilistic human pose modeling have employed normalizing flows to learn distributions parameterized by rotational measures \cite{xu_ghum_2020,kolotouros_probabilistic_2021}, they fall short in learning normalized densities on the SO(3). This is due to their Euclidean space-based approaches, which lack a continuous bijective mapping to SO(3).

To overcome these shortcomings, we introduce a normalizing flow defined on the product space of SO(3) manifolds,  accurately capturing the density of human joint rotations.
By defining flow layers that explicitly operate on SO(3), the learned density is normalized on the considered space of rotations with three degrees of freedom. The joint PDF on multiple SO(3) manifolds expresses the statistical dependencies between human joints accordingly. To achieve this, the SO(3) manifolds are explicitly linked using a nonlinear autoregressive conditioning not restricted to the common Gaussian setting.

We demonstrate that respecting the manifold structure of SO(3) in the normalizing flow design improves the performance for multiple applications that involve a probabilistic model of human pose, outperforming state-of-the-art approaches in most configurations.

In summary, our work makes three key contributions:

\begin{itemize}
    \item We present a normalizing flow model for learning normalized densities on the high-dimensional product space of SO(3) manifolds. This is enabled by conceptualizing a flow layer that specifically operates on SO(3) and lifting the model to a product space via an expressive nonlinear autoregressive conditioning approach. 
    \item Introducing HuProSO3, a probabilistic \textbf{Hu}man pose model on the \textbf{Pro}duct Space of \textbf{SO(3)} manifolds, we elucidate the application of our presented normalizing flow model in various applications where a probabilistic model of humans is relevant.
    \item We showcase HuProSO3's effectiveness and adaptability through applications like probabilistic inverse kinematics and 2D to 3D pose uplifting, outperforming several state-of-the-art methods in tasks involving probabilistic human pose models.
\end{itemize}

\section{Related work}
\label{sec:rel_work}
\subsection{Normalizing Flows on Rotational Manifolds}
Normalizing flows model complex distributions through a series of sequential flow layers. They learn a normalized density, represented by the change of variables formula:
\begin{equation}
  p(x)=\pi\left(T^{-1}(x)\right)\left| J_{T^{-1}(x)} \right|,
  \label{eq:nf}
\end{equation}
where $J_{T^{-1}(\cdot)}$ is the Jacobi-matrix of the inverse of the diffeomorphic transformation $T:\mathbb{R}^d\rightarrow\mathbb{R}^d$ and its inverse $T^{-1}(\cdot)$, where a diffeomorphic transformation is invertible and both, $T$ and its inverse $T^{-1}$,  are differentiable.

Several methods adapt normalizing flows, typically applied in Euclidean spaces, for rotational manifolds, including SO(3). \citet{pmlr-v119-rezende20a} successfully apply the Möbius transformation, circular splines, and a non-compact projection for learning densities on such manifolds. \citet{pmlr-v89-falorsi19a} develop a normalizing flow in Euclidean space that is then mapped to SO(3), but this approach involves non-unique mappings from the Lie algebra to the Lie group. Contrarily, \citet{liu_delving_2023} propose a novel Möbius coupling layer and quaternion affine transformation, facilitating the learning of normalizing flows directly on SO(3).

\textbf{Flows for Product Spaces of Rotational Manifolds.}
\citet{NIPS2017_6c1da886} present a strategy for an autoregressive model that allows learning densities on high-dimensional Euclidean spaces. Building on this, \citet{stimper_normflows_2023} introduced a PyTorch package for defining normalizing flows on Cartesian products of 1-spheres using two-dimensional embeddings of angular quantities. However, this method does not extend to general $n$-spheres or SO(3) manifolds. Additionally, \citet{glusenkamp_unifying_2023} facilitated learning on rotational manifolds, including tori and 2-spheres, using an autoregressive conditioning approach. Yet, his framework is limited by a fixed conditioning sequence and lacks support for flows on SO(3) manifolds.

\subsection{Learning Human Pose Distributions}
In the domain of human pose estimation, models traditionally use joint rotations as parameters \cite{akhter_pose-conditioned_2015, lehrmann_non-parametric_2013} to reflect the skeleton's rotational characteristics. Pioneering methods for learning the unconditional human pose distribution include a GMM \cite{leibe_keep_2016} and a VAE \cite{pavlakos_expressive_2019}, both proving effective as pose priors. Later advancements utilized normalizing flows in Euclidean space \cite{xu_ghum_2020} using the 6D representation \cite{zhou_continuity_2019}. 

Subsequent works \cite{davydov_adversarial_2022,avidan_pose-ndf_2022} that learn a human pose prior have highlighted the inadequacy of a Gaussian assumption for the joint rotation parameterization due to their unbounded nature. So, they propose different ways to account for this issue in the design of their generative models for the human pose. \citet{davydov_adversarial_2022} show that a spherical noise distribution in the latent space of GAN-based approach results in a distribution with more realistic human poses and a smoother latent space. 
\citet{avidan_pose-ndf_2022} model a manifold on the product space of SO(3) that represents the set of feasible human poses.

Probabilistic techniques have been effectively employed to infer human pose distributions from image inputs \cite{kolotouros_probabilistic_2021,sengupta2021hierprobhuman,sengupta_probabilistic_2021,sengupta_humaniflow_2023}. \citet{sengupta2021hierprobhuman} utilize the Matrix-Fisher distribution for learning the rotational distributions of individual joints on SO(3). Meanwhile, \citet{kolotouros_probabilistic_2021} model joint rotations using a 6D representation and employ normalizing flows to learn densities, incorporating an additional loss term to account for the orthonormality of the columns in the 6D representation.
\citet{sengupta_humaniflow_2023} present a novel approach to model human pose distributions on the product space of SO(3). This method factorizes the joint PDF by using autoregressive conditioning along the human kinematic tree. It respects the manifold structure of joint rotations by learning a normalizing flow on the Lie algebra and subsequently mapping it to the Lie group.
\citet{voleti_smpl-ik_2022} builds upon \cite{oreshkin2022protores:} to model the pose and applies it, e.g., to inverse kinematics from sparse 3D key points.

\section{Method}
\label{sec:method}
In this section, we describe our method for deriving and implementing the following desiderata: First, as outlined in \cref{sec:mcl}, each normalizing flow layer is designed to respect and operate within the SO(3) manifold. 
Second, we ensemble these individual flow layers to model a density on the product space of SO(3) manifolds.
Finally, we present the probabilistic human pose model HuProSO3, which comprises and instantiates the aforementioned methodology.

\subsection{Problem Statement}
Given the special orthogonal group SO(3) that is defined as the set of rotation matrices R
\vspace{-2pt}
\begin{equation}
	\{ \text{R} \in \mathbb{R}^{3\times 3}\,|\,\text{R}^T \text{R} = I,\quad \det \text{R} = 1\},
\vspace{-2pt}
\end{equation}
we aim to learn a PDF $p(\mathbf{R}|\mathbf{c})$, where the random variable \mbox{$\mathbf{R}=\left[\mathbf{R_1}, ..., \mathbf{R_N}\right]$} is defined on a product space $\mathbf{P}=\prod_i^N \mathcal{M}_i$ of $N$ SO(3) manifolds $\mathcal{M}_i = SO(3)$ and the context vector $\mathbf{c}\in\mathbb{R}^L$ conditions the PDF. 

\subsection{Normalizing Flows on a Product Space of SO(3) Manifolds}\label{sec:nf_cart_so3}
\begin{figure}[t] 
	\centering
	\includegraphics[width=\columnwidth]{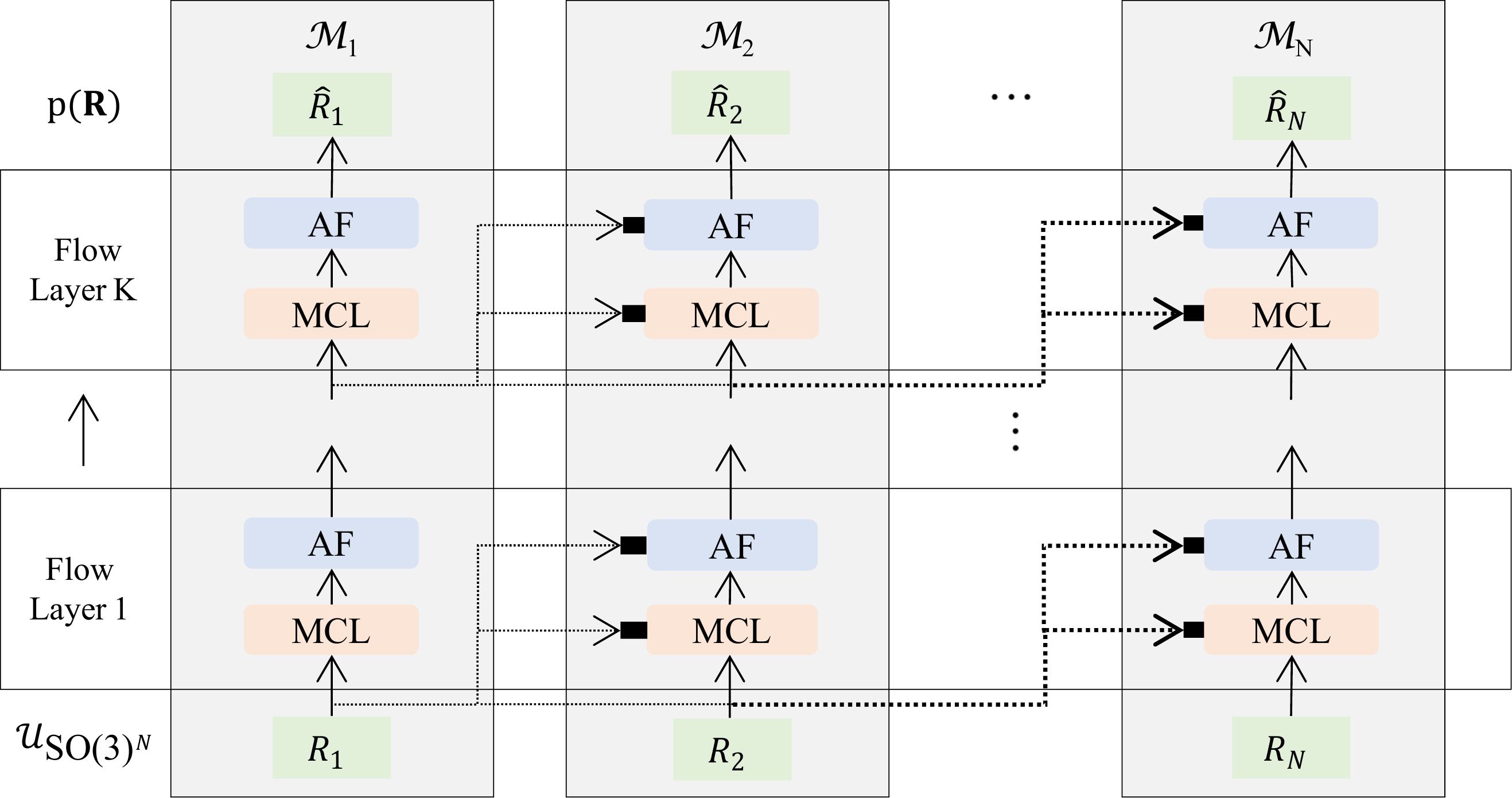}
	\caption{Overview of the components of the normalizing flow: The flow is defined on a product space of $N$ manifolds $\mathcal{M}_i=SO(3)$. It includes $K$ flow layers that transform samples $R_i$ from a uniform distribution on SO(3) to samples of the learned distribution $\hat{R}_j$. The flow is composed of a Möbius coupling layer (MCL) and a quaternion affine transformation (AF). 
 Vertical arrows indicate the flow of SO(3) samples through layers, while dotted arrows represent autoregressive conditioning using an MLP (black boxes).}
	\label{fig:nd_so3_schematic}
\vspace{-10pt}
\end{figure}

The structure of our model, as illustrated in \cref{fig:nd_so3_schematic}, involves multiple flow layers that convert a uniform density on SO(3) into the target density for each manifold $\mathcal{M}_i=SO(3)$. Samples $R_i$ are drawn from a uniform distribution on SO(3). 
Then, the samples are mapped to the target distribution $\hat{R}_i$ through the sequential application of diffeomorphic flow layers. The normalizing flow is trained by maximizing the likelihood of the target sample that is propagated through the flow layers (NLL Loss).

The manifolds are connected in an autoregressive manner to allow for learning a joint PDF. The flow is defined on SO(3) by leveraging a Möbius coupling layer and a quaternion affine transformation \cite{liu_delving_2023} as invertible flow layers on SO(3). The joint likelihood of the product space $\mathbf{P}$ is learned using randomly masked nonlinear autoregressive conditioning. We present these mechanisms in more detail in the following.

\textbf{Möbius Coupling Layers.} \label{sec:mcl}
The core principle of normalizing flows involves applying diffeomorphic transformations in each layer, which ensures that the original density remains normalized. However, defining these transformations for the SO(3) manifold, unlike in Euclidean space, presents challenges.
We must ensure that both the forward transformation and its inverse operate on SO(3) and do not map to points outside this manifold. To this end, we apply the Möbius transformation, which expresses parameterized distributions on spheres.

A rotation matrix $R \in \mathrm{SO}(3)$  can be described using any two of its three orthonormal three-dimensional unit column vectors $\left(\mathbf{x}_1 | \mathbf{x}_2 | \mathbf{x}_3 \right)$ \cite{zhou_continuity_2019}. Thus, each column represents a three-dimensional unit vector defined on the unit sphere $\mathbf{x}_i \in \mathcal{S}^2$, $i\in\{1,2,3\}$, parameterized in Euclidean coordinates $\mathbf{x}_i\in\mathbb{R}^3$.

The Möbius transformation of a point $\mathbf{y} \in \mathcal{S^D}$, given a parameter $\omega\in\mathbb{R}^{D+1}$, is defined as
\vspace{-2pt}
\begin{equation}
	h_\omega (\mathbf{y}) = h(\omega,\mathbf{y}) = \frac{1-||\omega||^2}{||\mathbf{y}-\omega||}.
 \vspace{-2pt}
\end{equation}
It projects a point $\mathbf{y}$ to a new location on the same sphere $\mathcal{S}^D$.
The Möbius transformation is leveraged to design the Möbius coupling layer~\cite{liu_delving_2023}---an invertible flow layer on SO(3) that transforms a valid rotation matrix \mbox{$R=\left(\mathbf{x}_1 | \mathbf{x}_2 | \mathbf{x}_3 \right)$} into a different valid rotation matrix 
\mbox{$R'=\left(\mathbf{x}_1' | \mathbf{x}_2' | \mathbf{x}_3' \right)$}. 
The Möbius coupling layer ensures orthogonality in the transformed rotation matrix as follows: The first column $\mathbf{x_1}'=\mathbf{x_1}$ remains unchanged, providing the basis for adapting the second column vector. This second column, modeled as a point $\mathbf{x}_2\in\mathcal{S}^2$ is transformed into a new point
	$\mathbf{x}_2' = h\left(\omega_\mathbf{x_1},\mathbf{x_2}\right) \in \mathcal{S}^2,$
using the Möbius transformation. The transformation parameter
\vspace{-2pt}
\begin{equation}
	\omega_\mathbf{x_1}=\xi(\mathbf{x_1})=g(\mathbf{x_1}) - \mathbf{x_1} \left(\mathbf{x_1}\cdot g(\mathbf{x_1})\right)
 \vspace{-2pt}
\end{equation}
is derived from $\mathbf{x_1}$.
Here, $g(\cdot)\in\mathbb{R}^3$ represents an MLP output, projected onto a plane perpendicular to $\mathbf{x}_1'$. Consequently, the resulting column vector $\mathbf{x}_2'$ is orthogonal to $\mathbf{x}_1'$. The final step in constructing the rotation matrix involves computing the third column, $\mathbf{x}_3'$, using the cross product $\mathbf{x}_3'=\mathbf{x}_1'\times\mathbf{x}_2'$. This ensures the completion of the rotation matrix $R'$ with a vector orthogonal to the first two columns.

In a rotation matrix, the first column specifies two out of three rotational degrees of freedom, while the second column determines the remaining degree. 
Since the first column $\mathbf{x}_1$ remains unchanged in a single Möbius coupling layer, each layer only modifies the rotation along one degree of freedom. 
To address this, we employ permutations in the conditioning sequence of consecutive flow layers. 
For example, we use the Möbius transformation to derive a new column $\mathbf{x}_1' = f\left(\omega_\mathbf{x_2},\mathbf{x_1}\right)$, where $\omega_\mathbf{x_2}$ is calculated from $\mathbf{x_2}$. 
By applying such permutations changes for several flow layers all rotational degrees of freedom can be modeled. 
These permutations, being invertible and volume-preserving transformations with an absolute Jacobian of 1 \cite{Papamakarios2019NormalizingFF}, can be integrated into normalizing flows.

While a single Möbius coupling layer effectively warps distributions on SO(3) nonlinearly, it only transforms one rotational degree of freedom via the second column of a rotation matrix. However, this strategy does not directly allow the modification of multiple rotational degrees simultaneously by moving or scaling distributions on SO(3), as \cite{liu_delving_2023} illustrates. To account for this, we also apply an affine quaternion transformation \cite{liu_delving_2023} in our normalizing flow.

\textbf{Quaternion Affine Transformation.}\label{sec:aft}
The quaternion affine transformation allows learning an affine transformation on the 3-sphere, a double-cover of SO(3). It, thus, realizes a global rotation and a shifting operation on SO(3).
The quaternion affine transformation 
\vspace{-2pt}
\begin{equation}
    f_\text{q}(\mathbf{q})=W\mathbf{q}\cdot (||W\mathbf{q}||)^{-1}
    \vspace{-2pt}
\end{equation}
is applied to a quaternion $\mathbf{q}=\text{m}_{\text{R}\rightarrow\mathbf{q}}(\textbf{R})$, where $\mathbf{q}$ is computed from the considered rotation matrix $\mathbf{R}$ and the invertible $4\times 4 $ matrix $W$ is constructed using \textit{svd} decomposition. The parameters of the matrices of the \textit{svd} decomposition are learned during training. As shown in \cref{fig:nd_so3_schematic}, this transformation follows each Möbius coupling layer.
Maintaining the quaternion's real part positive is required for the transformation's invertibility. This constraint does not hinder expressiveness, as $\mathbf{q}$ and $\mathbf{-q}$ denote the same rotation.

\textbf{Joint PDF on a Product Space.}
While the described Möbius and affine quaternion transformations effectively operate on a single SO(3) manifold, they do not suffice to construct a normalizing flow for the product space $\mathbf{P}=\prod_i^N SO(3)$.
We address this by introducing an autoregressive masking strategy, designed to learn a joint PDF on such a product space. 
Masked autoregressive flows \cite{Papamakarios2019NormalizingFF} learn PDFs defined on $n$-dimensional Euclidean spaces $\mathbf{z}\in\mathbb{R}^n$. They implement autoregressive conditioning by applying consecutive MADE~\cite{pmlr-v37-germain15} blocks for linking different dimensions with the Gaussian conditionals parameterized by
\vspace{-2pt}
\begin{equation}
\label{eq:masked_ar_flow}
    z_i = u_i \exp f_{\alpha_i}(\mathbf{z}_{1:i-1}) + f_{\mu_i}(\mathbf{z}_{1:i-1}),
    \vspace{-2pt}
\end{equation}
with the scalar functions $f_\cdot(\cdot)$ that output the mean and log standard deviation of the conditional $i$ given all previous dimensions. The term $u_i\sim \mathcal{N}(0,1)$  represents noise.

The expression in \cref{eq:masked_ar_flow} links individual dimensions. However, this approach does not account for manifolds that cannot be expressed by Cartesian products in Euclidean space, e.g., the 2-sphere or SO(3). 
Using standard Gaussian conditionals falls short in capturing the dependencies of non-Euclidean manifolds.

To address these challenges, we first apply autoregressive conditioning based on random variables defined on SO(3). For this purpose, the joint density is decomposed autoregressively: 
\vspace{-2pt}
\begin{equation}
	p(\mathbf{R}) = \prod_i^N p(\mathbf{R}_i | \mathbf{R}_{1:i-1}),
    \vspace{-2pt}
\end{equation}
where each conditional is defined on $\mathbf{R}_i\in\text{SO}(3)$ and conditioned on preceding SO(3) manifolds $\mathbf{R}_{1:i-1}\in\prod_1^{i-1}\text{SO}(3)$.
Such a decomposition is generally restricted to a fixed order of conditioning, which does not capture arbitrary dependencies between different SO(3) manifolds. To mitigate this problem, we randomly vary the order of conditioning in each flow layer, similar to \cite{NIPS2017_6c1da886}, which results in more expressive flows \cite{Papamakarios2019NormalizingFF}.
We achieve this by permuting the SO(3) rotation sequences between autoregressive layers, leveraging the invertibility of permutations.

Instead of using a linear coupling with the conditionals parameterized as Gaussians for Euclidean space as proposed in \cite{NIPS2017_6c1da886} and expressed in \cref{eq:masked_ar_flow}, we compute the parameters of the considered manifold $\mathcal{M}_i$ autoregressively. This involves a nonlinear map $g_\text{c}\left( \mathbf{x}_{12}^{(1:i-1)} \right)$  with parameters learned during training to condition the current manifold's distribution on the preceding ones. We parameterize the SO(3) rotations of previous dimensions $\mathbf{R}_{1:i-1}$ in the continuous 6D representation~\cite{zhou_continuity_2019} with $\mathbf{x}_{12}^{(i:i-1)}$.
This conditioning is then integrated into both the affine transformation and the Möbius coupling layer.

The affine transformation is conditioned on the previous dimensions by using a neural network to compute the matrix $\mathbf{W}_i=g_\text{c-W}^{(i)}\left( \mathbf{x}_{12}^{(1:i-1)} \right)$.
We realize the autoregressive conditioning in the Möbius coupling layer by computing the parameters of the Möbius transformation based on the first rotation matrix' column of the considered dimension $\mathbf{x}_1^{(i)}\in\mathbb{R}^3$ and the previous dimensions $\mathbf{x}_{12}^{(1:i-1)}\in\mathbb{R}^{3\times2\times(i-1)}$ in 6D representation $g_\text{c-M}\left(\left[ \mathbf{x}_1^{(i)} ,  \mathbf{x}_{12}^{(1:i-1)} \right]\right)$.

\textbf{Conditioning the Joint PDF.}
\label{par:cond_pdf}
The joint PDF can be conditioned on additional information $\mathbf{c}$ by augmenting the set of conditions of each term in the chain rule of probability with the condition $\mathbf{c}$:
\vspace{-2pt}
\begin{equation}
    \label{eq:cond_flow}
	p(\mathbf{R}|\mathbf{c}) = \prod_i p(\mathbf{R}_i | \mathbf{R}_{1:i-1},\mathbf{c}).
\end{equation}

\subsection{Learning Human Pose Distributions}
\label{sec:huproso3}
We leverage our normalizing flow to learn a probabilistic model of the human pose. We use the SMPL~\cite{leibe_keep_2016} model to parameterize the human pose with $21$ joints. After removing static joints in the AMASS database, our model describes a density $p(\mathbf{R}|\mathbf{c})$ on a product space of $N=19$ SO(3) manifolds. 
We call this normalizing flow HuProSO3, a probabilistic \textbf{Hu}man pose model on the \textbf{Pro}duct Space of \textbf{SO(3)} manifolds.

We deploy HuProSO3 in several applications, the first being to learn an unconditional human pose prior. In this case, we do not condition, i.e., $\mathbf{c}=\emptyset$. This pose prior acts as a generative model for sampling realistic human poses and can also be used for evaluating the probability of a specific pose that is parameterized by joint rotations $p\left(\mathbf{R}=\{R_\text{joint,1}, ...,R_\text{joint,19}\}\right)$. 
Second, we inject a condition $\mathbf{c_\text{feat}}$ into the model to learn a density that is specific to a given context. To handle varying contexts and to remove computations in each flow layer, we add an additional MLP $\mathbf{c}=g(\mathbf{c}_\text{feat})$  that computes the relevant features that are injected into the normalizing flow, as outlined in \cref{eq:cond_flow}.
HuProSO3 generally supports arbitrary conditions. We present conditioning with 2D and 3D key points, where $\mathbf{c}_\text{feat}\in\mathbb{R}^{J\times k}$ with $J=21$ joints and $k\in\{2,3\}$ for conditioning with 2D and 3D key points.

To demonstrate HuProSO3's ability to capture human pose distributions, we adapt the masking strategy from \cite{ci_gfpose_2022}. Parts of the context features $\mathbf{c}_\text{feat}'=\mathbf{m}\cdot\mathbf{c}_\text{feat}$ are removed by applying the mask $\mathbf{m}=\left[m_1, m_2, ..., m_J\right]$ with $m_i\in\{0,1\}$ indicating whether a joint position is accessible. To allow conditioning with an arbitrary number of joints during evaluation, we seek a different masking strategy to \cite{ci_gfpose_2022} and we mask each joint with varying probabilities $p_m \in \left[ 0,1 \right]$ during training. Therefore, the model has learned to handle varying numbers of occluded joints.

\section{Experiments}
\label{sec:experiments}
While HuProSO3's design ensures that all flow transformations are on SO(3) and capture nonlinear dependencies between the SO(3) manifolds, we now also demonstrate experimentally that it has sufficient expressiveness to accurately model human pose distributions. 
We, therefore, evaluate HuProSO3 on different applications that involve a probabilistic model of the human pose.
First, we learn an unconditional human pose prior that showcases HuProSO3's capability of learning the intricate distribution human poses. 

Second, we condition the distribution demonstrating HuProSO3's capabilities of injecting observations or other conditions for learning task-specific distributions and correctly adapting the respective uncertainties arising from different sources of conditioning. For this task, we also evaluate cases when only partial information is given. This includes scenarios with partial information, such as occluded joints, where HuProSO3 effectively reasons about uncertainties due to missing information.

\begin{table}
\caption{Summary of precision and recall statistics for AMASS. The reported values represent the cumulative geodesic distances for all joint rotations between samples from the dataset and the evaluated pose prior.}
    \centering
    \footnotesize
    \begin{tabular}{@{}lcc|ccc}
        \toprule
        & \multicolumn{2}{c}{Test (mean [median])} & \multicolumn{2}{c}{Train (mean [median])} \\
        \midrule
        & Recall & Precision & Recall & Precision \\
        \midrule 
        GAN-S \cite{davydov_adversarial_2022} & 3.76 [3.34] & 4.51 [4.23] & 3.57 [3.34] & 4.38 [4.13] \\
        6D NF & 3.66 [3.16] & 4.50 [4.00]  & 3.55 [3.32] & 4.42 [4.10] \\
        Ours & \textbf{3.44 [2.95]} & \textbf{4.24 [3.71]}  & \textbf{2.93 [2.64]} & \textbf{3.90 [3.59]} \\
        \bottomrule
    \end{tabular}
    \label{tab:combined_pr_results}
    \vspace{-10pt}
\end{table}

\subsection{Unconditional Pose Prior} 
To demonstrate HuProSO3's effectiveness in an unconditional setting, we train a human pose prior using the AMASS database \cite{mahmoodAMASSArchiveMotion2019} with its standard dataset split. We then compare its ability to generate realistic human poses against various other human pose priors.

\textbf{Metrics.}
In general, the likelihood is the best metric for evaluating how well a model has learned a distribution. However, we are the first to provide normalized joint densities of the human pose distribution on SO(3) manifolds. Therefore, our evaluation relies on comparing generated samples from HuProSO3 with those from previous approaches. We follow an experimental pipeline similar to \cite{davydov_adversarial_2022} to evaluate whether human poses sampled from the considered pose prior deviate from samples from the dataset distribution (precision) and whether the pose prior has captured the variety of poses in the dataset (recall). For recall, we compare $1$k dataset samples against $100$k poses sampled from our model, calculating the minimum error (nearest neighbor). Precision is evaluated by comparing $100$k dataset samples against $1$k model-generated samples, again computing the minimum error for these model samples.

Since we aim to evaluate how well the model captures the distribution $p(\mathbf{R})$ on the product space of SO(3), we evaluate precision and recall based on the sum of all $J=21$ geodesic distances $d_\text{geo}(R_{i,k},R_{j,k})$ of the rotations $R_{i,k}$ and $R_{j,k}$ that correspond to the $k\text{-th}$ SMPL joint of pose $i$ and $j$, respectively,
where the geodesic distance is computed by
\begin{equation}
    d_\text{geo}(R_{i,k},R_{j,k})=\arccos{\frac{\text{tr }(R_{i,k}^TR_{j,k})-1}{2}}.
\end{equation}

\textbf{Baselines.}
In our evaluation, HuProSO3 is compared with the GAN-S human pose prior \cite{davydov_adversarial_2022} and a 6D normalizing flow similar to models in \cite{xu_ghum_2020, vedaldi_weakly_2020}. We train GAN-S from scratch using the standard parameter configuration provided by \cite{davydov_adversarial_2022}. To compare against a 6D normalizing flow, we trained an autoregressive \cite{NIPS2017_6c1da886} neural spline flow \cite{NEURIPS2019_7ac71d43} using the normflows library \cite{stimper_normflows_2023}. We provide additional results for Pose-NDF \cite{avidan_pose-ndf_2022} as a pose prior in see \cref{tab:pr_results_posendf} (\cref{sec:eval_ll}).

\textbf{Results.} In \cref{tab:combined_pr_results}, we present the mean and median of precision and recall for both, training and test datasets. The results illustrate that HuProSO3 has captured the training dataset's distribution best. The means of precision and recall metric, both, are lower depicting that the model generates poses that correspond to realistic poses from the dataset but it also captures the variety of the seen poses during training. We observe similar results for the median metric.
Moreover, while having learned the training distribution more accurately, our model does not overfit and it outperforms other methods on the precision and recall metric for the test dataset as well. Precision and recall curves, log likelihood comparisons for training and test datasets, and qualitative joint correlation illustrations are provided in the appendix. Additionally, rendered poses in \cref{fig:humans_rendered} confirm HuProSO3's ability to generate realistic and diverse human poses (see \cref{sec:samples_prior}).

\begin{table}
\caption{MGEO [rad] and MPJPE [mm] results for the IK task on the AMASS test split. Performance comparisons are made for optimization-based (\underline{best}) and probabilistic methods (\textbf{best}). Probabilistic methods are assessed using a single random sample and the average of 10 random samples.}
    \centering
    \footnotesize
    \begin{tabular}{@{}lcc}
        \toprule
         Method&  MPJPE (1 $|$ 10)&  MGEO (1 $|$ 10)\\
         \midrule
         VPoser \cite{pavlakos_expressive_2019}&  26.6 $|$  - &  \underline{0.230} $|$  -\\
        GAN-S \cite{davydov_adversarial_2022}& 12.1$|$  -&0.224 $|$  -\\
        Pose-NDF \cite{avidan_pose-ndf_2022}&  \underline{0.2} $|$  -& 0.289 $|$  -\\
        \midrule
        HF-AC \cite{sengupta_humaniflow_2023}& 33.2 $|$  21.2& 0.308 $|$  0.251\\
         Ours&  8.2 $|$  \textbf{5.5}&  0.163 $|$  \textbf{0.121}\\
         \bottomrule
    \end{tabular}
    \label{tab:eval_inv_kin}
\vspace{-10pt}
\end{table}

\subsection{Inverse Kinematics}
\label{sec:exp_ik}
In this experiment, we condition the learned density on the 3D position of the $J=21$ SMPL joints to solve the human pose inverse kinematic (IK) problem: The goal is to learn the PDF  $p\left(\textbf{R}\vert \mathbf{c}\right)$ that is defined on SO(3) and conditioned on the joint positions in 3D space $\mathbf{c}\in\mathbb{R}^{J\times 3}$.

\textbf{Metrics.}
To assess the models, we use the Mean Geodesic Distance (MGEO) across joint rotations for rotational error and the Mean Per Joint Position Error (MPJPE) to evaluate the accuracy of learned densities in relation to error propagation in forward kinematics. Effective probabilistic models are expected to produce samples consistent with the given conditioning, with accuracy increasing upon evaluating more samples. Therefore, we test probabilistic methods using both a single random sample and an average of 10 random samples. 

\textbf{Baselines.}
We compare our approach against both, optimization-based and learning-based baselines by evaluating the MGEO and MPJPE metrics. We consider the following learned pose priors: We use the IK solver provided by \cite{pavlakos_expressive_2019} focusing solely on pose without shape optimization. Furthermore, we evaluate PoseNDF \cite{avidan_pose-ndf_2022} utilizing the pre-trained model and the existing pipeline. 
As implemented in the published code by \cite{avidan_pose-ndf_2022}, we use the optimization objective
\vspace{-3pt}
\begin{equation}
\label{eq:loss_ik}
\mathcal{L}_\text{IK}=\sum_{k=1}^{N_\text{test}}\sum_{i=1}^{J}\left||\textbf{x}_{\text{gt},i,k}-\hat{\textbf{x}}_{i,k}\right||_2,
\vspace{-2pt}
\end{equation}
which compares the ground truth joint position $i$ $\mathbf{x}_{\text{gt},i,k}$ with the predicted position $\hat{\textbf{x}}_{i,k}=FK(\mathbf{R}_{i,k})$ after applying forward kinematics to the optimized rotation and all $N_\text{test}$ samples of the test dataset. We utilize the optimization process from~\cite{avidan_pose-ndf_2022}, omitting the temporal smoothness term.

Similarly, we use the trained GAN-S model and we optimize its latent code $\textbf{z}$ to minimize the loss in \cref{eq:loss_ik} with $\hat{\textbf{x}}_{i}=\mathcal{G}(\mathbf{z})$ using the LBFGS optimizer~\cite{liu_limited_1989}. 

In addition, we use the implementation of HuManiFlow~\cite{sengupta_humaniflow_2023}, which implements the normalizing flow for learning the ancestor-conditioned density on a single SO(3) manifold. Different from its original design, which conditions on visual features from a CNN encoder, we adapt it to condition on 3D keypoint positions and train the normalizing flow from scratch. We refer to it with HF-AC.

\textbf{Results.}
The low errors, as depicted in \cref{tab:eval_inv_kin}, show that HuProSO3 is capable of respecting conditions accurately. Our probabilistic method has mostly comparable or better results than other pose priors optimized in latent space \cite{davydov_adversarial_2022,pavlakos_expressive_2019}.
Pose-NDF performs best on the MPJPE metric because it does not penalize a pose as long as it is realistic. For this reason, the optimization-based approach allows to match the given 3D key points nearly perfectly. However, the worse performance on the MGEO metric demonstrates that Pose-NDF models the manifold of plausible poses. 
Therefore, it cannot infer the most likely joint rotations for the given 3D key points, but it infers a rotation that belongs to the set of feasible joint rotations.

We evaluate the exact log-likelihood of HuProSo3 and HF-AC approach in \cref{tab:ll_evals} in \cref{sec:eval_ll}.

\begin{table}
\caption{MPJPE $|$ MGEO for different types of occlusions: Left leg (L), left arm and hand (A+H), and right shoulder and upper arm (S+UA). 
}
    \centering
    \footnotesize
    \begin{tabular}{@{}lccc}
        \toprule
        Method & L& A+H & S+UA\\
        \midrule
        Pose-NDF \cite{avidan_pose-ndf_2022}  & 28.3 $|$ 0.341 & 38.3 $|$ 0.360 & \underline{2.1} $|$ 0.333\\
        GAN-S  \cite{davydov_adversarial_2022}    & \underline{16.8} $|$ \underline{0.240}& \underline{29.6} $|$ \underline{0.260}& 18.8 $|$ \underline{0.241}\\
        \midrule
        HF-AC \cite{sengupta_humaniflow_2023} (N=10) & 101.3 $|$ 0.289 & 81.4 $|$ 0.301 & 72.2 $|$ 0.270 \\
        Ours (N=10)  & \textbf{13.9} $|$ \textbf{0.165}& \textbf{30.2} $|$ \textbf{0.201}& \textbf{12.7} $|$ \textbf{0.162}\\
        \bottomrule
    \end{tabular}
    \label{tab:partial_ik_mgeo_mjpjpe}
    \vspace{-10pt}
\end{table}

\subsection{Inverse Kinematics with Partial Observation}
We evaluate the capabilities of HuProSO3 to solve IK in the case of partially given 3D key points, e.g., when some joints are occluded. 
The rotational distribution of these joints can be inferred using a probabilistic human pose model.
A joint PDF is essential for this task because it captures all statistical dependencies, which contrasts selecting a fixed sequence of conditioning, e.g., along the kinematic tree. We provide a dataset analysis in \cref{sec:dataset_analysis}.
Following the methodology of \cite{avidan_pose-ndf_2022, rempe_humor_2021}, we evaluate three examples of occlusions: the left leg (L), the left arm including the hand (A + H), and the right shoulder with the upper arm (S + UA).

We utilize the same HuProSO3 model across all experiments, conditioned on 3D keypoint positions and trained with the randomized masking strategy outlined in \cref{sec:huproso3} to account for occluded joints. This approach enables the model to handle various occlusion scenarios effectively.

\textbf{Metrics and Baselines.}
Similar to the IK setting elaborated above, we report MGEO and MPJPE and we compare with Pose-NDF and GAN-S, as detailed in \cref{sec:exp_ik}.
Additionally, we compare HuProSO3 to previously reported results based on the per-vertex error as in \cite{rempe_humor_2021} and \cite{avidan_pose-ndf_2022} (see \cref{tab:pve_occ} in \cref{subsec:per_vertex_error_ik}).

To account for the occluded joints, we mask the occluded joints in the optimization objective similar to the conditioning for HuProSO3 in \cref{sec:huproso3} and we augment the optimization objective \cref{eq:loss_ik}
\vspace{-2pt}
\begin{equation}
\label{eq:loss_masked_ik}
\mathcal{L}_\text{IK}=\sum_k^{N_\text{test}}\sum_i^{J}m_i\left||\textbf{x}_{\text{gt},i,k}-\hat{\textbf{x}}_{i,k}\right||_2,
\vspace{-2pt}
\end{equation}
with the mask \mbox{$\mathbf{m}=\left[m_1,...,m_J\right],\  m_i\in\{0,1\}$}, that cancels the contributions of the occluded joints to the loss term. 
The joint rotations are initialized randomly but close to 0, as stated in \cite{avidan_pose-ndf_2022}.
\begin{figure}[t] 
	\centering
	\includegraphics[width=1.\columnwidth]{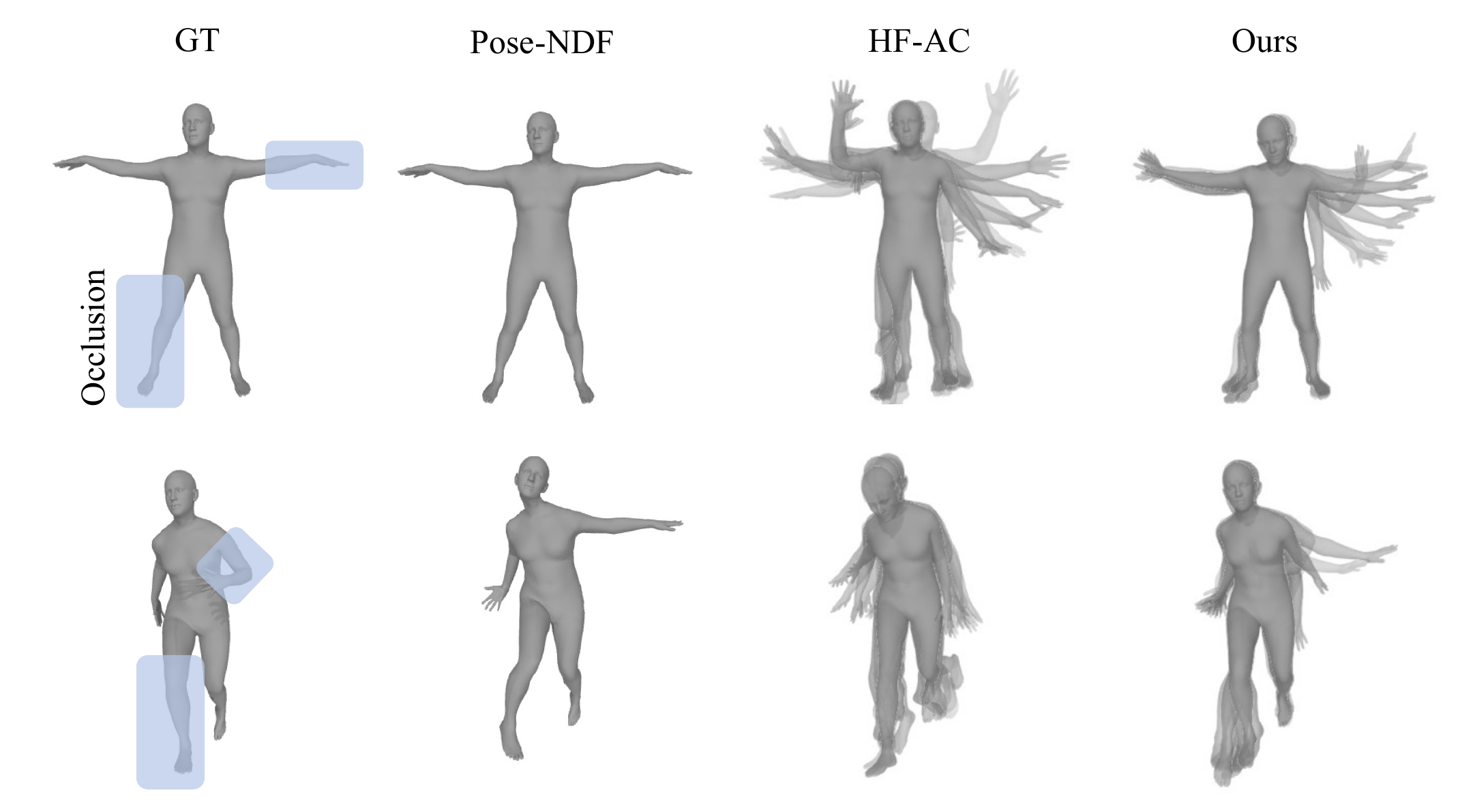}
 \vspace{-10pt}
	\caption{Qualitative results for inverse kinematics with partial occlusion (left arm and right leg). 
 For the normalizing flow based models (HF AC and HuProSO3), we visualize 10 samples, where less likely poses are more transparent.
 }
	\label{fig:qualitative_partial_ik}
\vspace{-5pt}
\end{figure}

\textbf{Results.}
We present the MGEO and MPJPE evaluation results in \cref{tab:partial_ik_mgeo_mjpjpe}. 
Optimization-based methods Pose-NDF \cite{avidan_pose-ndf_2022} and VPoser \cite{pavlakos_expressive_2019} are initialized with rotations close to the rest pose. 
For this reason, they perform significantly better for occluded legs, where the ground truth involves mostly straight legs.
However, these methods yield higher errors when optimizing the more variable arm joints. In contrast, HuProSO3 demonstrates more accurate estimates across various types of occlusions. 
The results show HuProSO3 is excelling for all occlusion types on the MGEO metric. It better captures the distribution on the joint rotation space.
We present the per-vertex errors in \cref{tab:pve_occ} in \cref{subsec:per_vertex_error_ik}.

Pose-NDF performs exceptionally well in cases of right shoulder and upper arm occlusions, likely due to the accurate optimization of arm joint rotations based on the given hand position. 
However, the MGEO metric is comparably high. As elucidated for the IK task, Pose-NDF does not represent a probabilistic model of the human pose but models the manifold of feasible poses. 
For this reason, the most likely joint rotation is not inferred, but only one possible rotation is computed. 
Therefore, large errors arise in particular for the rotations of the leaf joints (see, e.g., the right hand in the second row of \cref{fig:qualitative_partial_ik}).

\begin{figure}[t] 
	\centering
	\includegraphics[width=0.8\columnwidth]{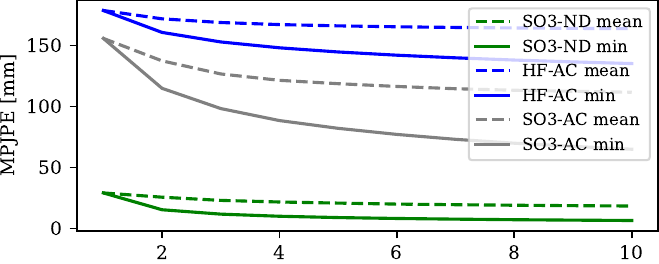} 
	\caption{Minimum MPJPE and MPJPE of the mean pose for HuProSO3, ancestor-conditioned SO(3), and HF-AC for randomly occluded joints ($p_m=0.3$) with varying numbers of samples.}
	\label{fig:ablation}
 \vspace{-5pt}
\end{figure}

\begin{table}
\caption{Evaluation results for the 5-point evaluation task in MPJPE [mm] and MGEO [rad]. We consider 1 sample and the average over 10 samples for HuProSO3.}
    \centering
    \footnotesize
    \begin{tabular}{@{}lcc|c}
        \toprule
         Method&  Pose-NDF \cite{avidan_pose-ndf_2022}&  GAN-S \cite{davydov_adversarial_2022} & Ours (1 $|$ 10) \\
         \midrule
         MPJPE & \underline{37.8}&46.5 & 36.4 $|$  \textbf{27.3} \\
         MGEO & 0.405&\underline{0.284}&0.290 $|$ \textbf{0.208} \\
         \bottomrule
    \end{tabular}
    \label{tab:AMASS_end_effector}

\vspace{-5pt}
\end{table}

GAN-S shows a lower MPJPE for occlusions of the left arm and hand, with performance comparable to HuProSO3 for other occlusions. This indicates the effectiveness of the latent code optimization in GAN-S, especially on the MPJPE metric. 
However, our model captures the joint rotation distribution more accurately, as the better MGEO metric depicts. In addition, our model predicts without the need for optimization on the joint positions, without being specifically trained for that particular type of occlusion, and it predicts in a probabilistic manner.
To demonstrate how a varying number of randomly occluded joints affects performance, we refer to \cref{fig:vary_pmask} in \cref{subsec:vary_number_samples}.

For the scenario of occluded joints, HuProSO3 outperforms HF-AC on rotation- and position-based metrics. We reason that this is, firstly, due to HuProSO3 modeling the complete joint density while HuManiFlow~\cite{sengupta_humaniflow_2023} learns only the ancestor-conditioned distribution of individual joints. 
If we train our model with a fixed ancestor-conditioning, we still reach a better accuracy than HF-AC for randomly occluded joints (\cref{fig:ablation}). See more details in \cref{subsec:vary_number_samples}.
We assume that designing the normalizing flow directly on SO(3) contributes to the better performance. 

\cref{fig:qualitative_partial_ik} illustrates qualitative results for IK with occluded joints.
HuProSO3 accounts for the ambiguous nature of the task: While visible joints only have a low diversity, occluded joint result in diverse but likely predictions.
We compare the diversity of the predictions to the model extracted from \cite{sengupta_humaniflow_2023} as proposed by \cite{sengupta_humaniflow_2023} by computing the averaged Euclidean deviation from the sample mean for all generated samples. We evaluate for $10$k random samples from the AMASS test datasets. Here, our model has a sample diversity of 11\ mm and 109\ mm for visible and occluded joints, respectively, while the ancestor-conditioned model by \cite{sengupta_humaniflow_2023} results in a diversity of 39\ mm and 98\ mm for visible and occluded joints, respectively.

\textbf{5-Point Evaluation.}
We also assess HuProSO3 on the 5-point evaluation benchmark \cite{oreshkin2022protores:,voleti_smpl-ik_2022}, where the model must infer all joint rotations based solely on the leaf joints. 
Despite not being explicitly trained for this specific condition, HuProSO3 successfully infers the pose distribution from the 3D keypoint positions of the leaf joints (see \cref{tab:AMASS_end_effector}).
\begin{figure}[t] 
	\centering
	\includegraphics[width=\columnwidth]{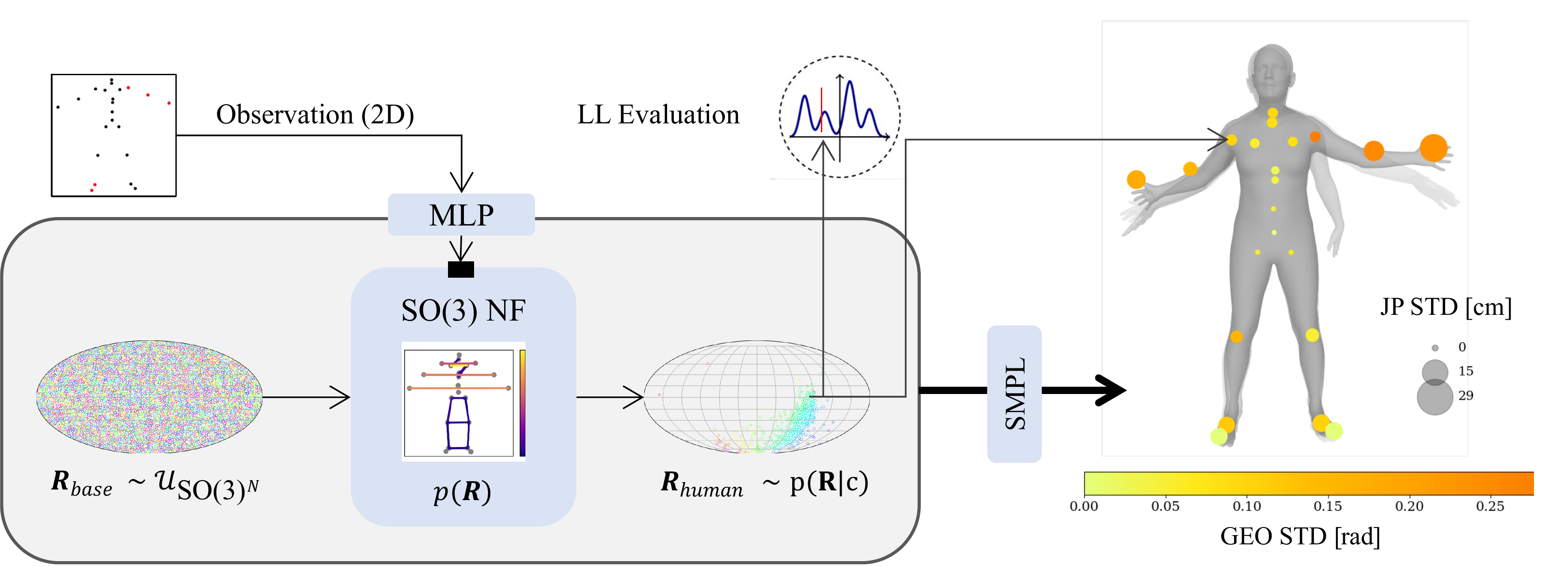} 
	\caption{Overview of an example application of HuProSO3: Samples from $p(\mathbf{R}|\mathbf{c})$ are generated by propagating samples from a uniform distribution on the product space of SO(3) through the normalizing flow. Partially given 2D key points serve as conditioning, where occluded joints are depicted in red.
 The prior $p(\mathbf{R})$ captures the statistical dependencies between different joints. 
 The color of the blobs depict the standard deviation of the joint rotations computed for 20 samples. The blob size relates to the standard deviation of the joint position (JP) after applying forward kinematics. 
 }
	\label{fig:huproso_2d}
 \vspace{-5pt}
\end{figure}

\subsection{2D to 3D Uplifting via SO(3) Parameterization}
In this experiment, we condition the learned density with 2D keypoint positions $\mathbf{x}_\text{2D}\in\mathbb{R}^{J\times 2}$ of the $J=21$ SMPL joints and learn $p\left(\textbf{R}\vert \mathbf{c}\right)$, where  $\mathbf{c}=g_\text{feat}(\mathbf{x}_\text{2D})$ is computed by an MLP. Then, the 3D pose is obtained by applying forward kinematics. We illustrate this setting in \cref{fig:huproso_2d}. 

\textbf{Metrics and Baselines.}
For the 2D to 3D uplifting application, we use MGEO and MPJPE as evaluation metrics, similar to the IK task. Our model is compared with Pose-NDF, GAN-S, and HF-AC. Unlike the previous experiments focusing on 3D keypoint positions (\cref{eq:loss_ik}), here we optimize for 2D keypoint positions.

\textbf{Results.} 
Evaluation results for the 2D to 3D uplifting task are detailed in \cref{tab:amass_2d_so3}, focusing on models using rotation representations for human pose. Unlike the IK task, inferring human poses from 2D data involves significant ambiguity. 
The results show that the considered optimization-based methods are performing worse since the optimization objective is less expressive and the results depend on the initialization. 
Our model HuProSO3 outperforms other methods, affirming its effectiveness in probabilistically capturing ambiguous input conditions.

\cref{fig:huproso_2d} qualitatively illustrates HuProSO3's applications for partially given 2D key points. The generated samples depict the arising uncertainties accordingly. More samples are visualized in \cref{fig:humans_rendered_uplifting} in \cref{sec:samples_2d_to_3d}.

\begin{table}
\caption{MPJPE [mm] and MGEO [rad] results for the 2D to 3D uplifting task on the AMASS test dataset. For probabilistic models, results are shown for both a single sample and the average of 10 samples.}
    \centering
    \footnotesize
    \begin{tabular}{@{}lcc}
    \toprule
     Method&  MPJPE (1 $|$ 10)&MGEO (1 $|$ 10)\\
     \midrule
    Pose-NDF \cite{avidan_pose-ndf_2022}& 133.1 $|$  -&0.522 $|$  -\\
    GAN-S \cite{davydov_adversarial_2022}& \underline{61.5} $|$  -&\underline{0.277} $|$  -\\
     \midrule
     HF-AC \cite{sengupta_humaniflow_2023}&  74.4 / 56.0& 0.346 / 0.278\\
     Ours& \textbf{32.8} $|$  \textbf{23.4}&\textbf{0.209}  $|$  \textbf{0.153}\\
     \bottomrule
    \end{tabular}
    \label{tab:amass_2d_so3}
    \vspace{-5pt}
\end{table}

\section{Conclusion}
\label{sec:conclusion}
We addressed the gap in normalized density models for high-dimensional product spaces of rotational manifolds, crucial in human pose modeling, which depends on rotational quantities. 
Our solution, HuProSO3, is a normalizing flow that effectively learns densities on product spaces of SO(3) manifolds, capturing the rotational nature of human poses in a probabilistic way. 
Demonstrating its efficacy, HuProSO3 not only excels as an unconditional pose prior in generative applications but also adapts to applications that require conditioning on potentially incomplete information, as illustrated for the task of inverse kinematics and 2D to 3D uplifting via a rotational SO(3) parameterization.

\textbf{Limitations and Future Work.} While effective in practice, the autoregressive structure in HuProSO3 has limitations in capturing all dependencies in a high-dimensional space, as it heavily depends on the permutation operation of the conditioning sequence. 
Moreover, sampling from a high-dimensional autoregressive model is slow as it scales linearly with the number of manifolds.
Another limitation is that our model currently only supports products of SO(3) manifolds. 
To better reflect the biomechanical structure of different joints, our approach could be extended to include other rotational manifolds. 
Additionally, the ability of our method to compute normalized densities opens up applications in human-robot collaboration. 
It also allows the integration as a prior or the injection of measurements in filtering and estimation problems, e.g., for human pose estimation based on key point measurements of the joints.

\textbf{Acknowledgments.} The authors would like to thank the Ministry of Science, Research and Arts of the Federal State of Baden-Württemberg for the financial support of the project within the InnovationCampus Future Mobility (ICM).

\clearpage
{
    \small
    \bibliographystyle{ieeenat_fullname}
    \bibliography{main}
}

\clearpage
\appendix
\setcounter{page}{1}
\maketitlesupplementary

\noindent To complement the main paper, this supplementary material delves into additional aspects not covered in detail previously. We include an analysis of the AMASS database, provide extended qualitative and quantitative evaluations, and share details on the implementation and training.

\section{Dataset Analysis} \label{sec:dataset_analysis}
Given that both the development and assessment of our method are grounded in a statistical analysis of the AMASS database, we present the key findings and insights from this analysis in this section.
\begin{figure}
    \centering
    \includegraphics[width=\columnwidth]{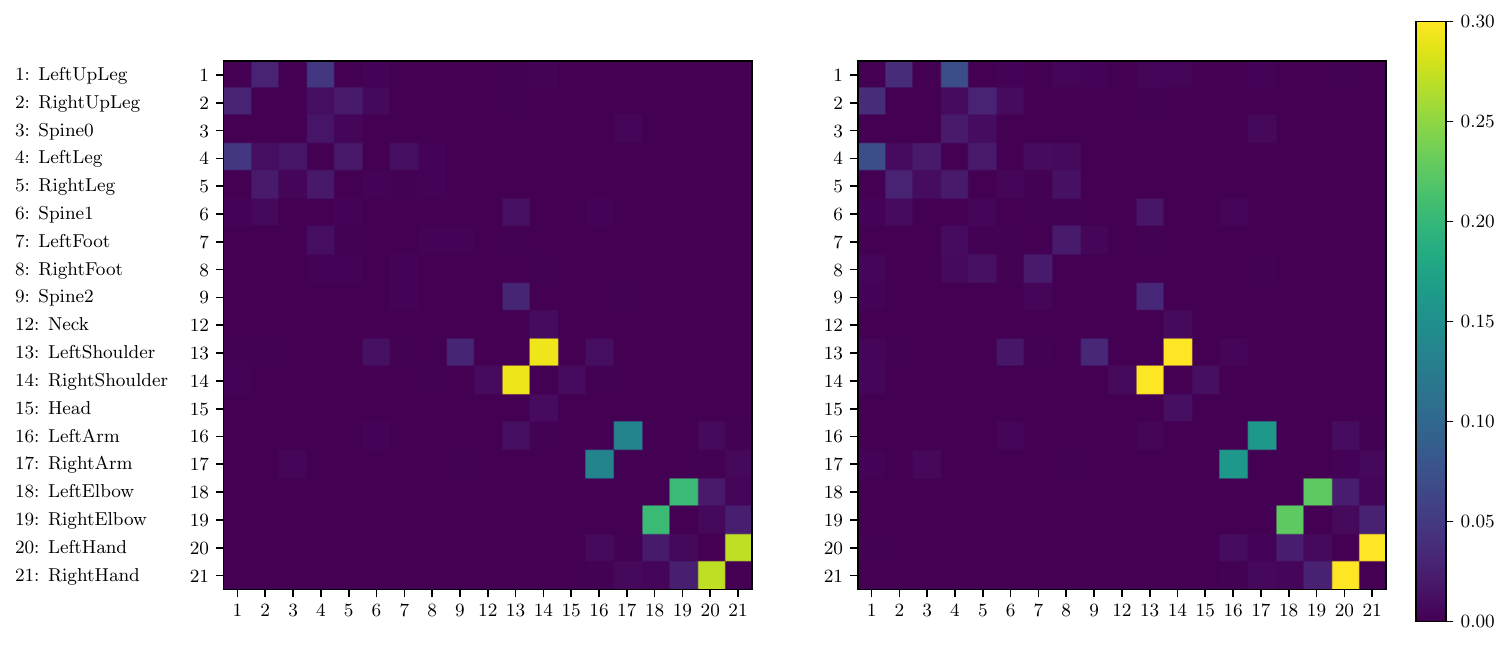}
    \caption{Spherical correlation coefficient of samples from HuProSO3 (left) and from AMASS training dataset (right), excluding toe bases and root joints. Entries on the diagonal are set to zero because their correlation coefficient $\hat{\rho}_{X Y}=1$ is always equal to one. The coefficients are computed using $100$k samples.}
    \label{fig:cc_joints}
    \vspace{-10pt}
\end{figure}

\subsection{Correlation Between Different Joints}
\begin{figure}
    \centering
      \begin{subfigure}{0.45\columnwidth}
        \includegraphics[width=0.9\linewidth]{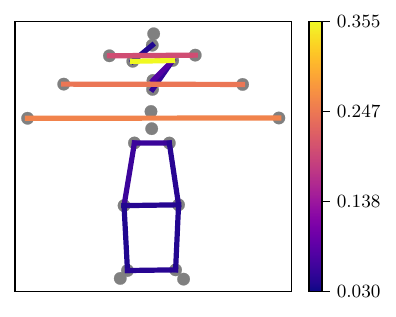}
        \caption{All coefficients.}
        \label{fig:cc_kin_all}
      \end{subfigure}
      \begin{subfigure}{0.45\columnwidth}
        \includegraphics[width=0.9\linewidth]{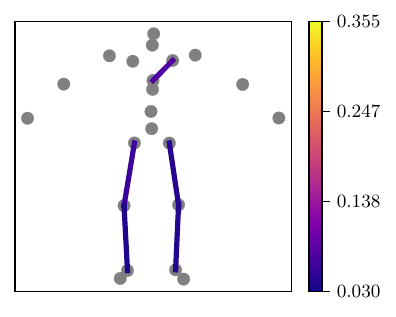}
        \caption{Along kinematic tree.}
        \label{fig:cc_kin_kin}
      \end{subfigure}
    \caption{Color-coded spherical correlation coefficients illustrated along a skeleton model of the human body. For clarity, only the coefficients with a value $\hat{\rho}_{X Y}>0.03$ are depicted.}
    \label{fig:cc_kin}
    \vspace{-10pt}
\end{figure}

Computing a correlation or dependence coefficient on SO(3) is not straightforward. Hence, we use a unit quaternion representation of orientations and follow \cite{fisher_correlation_1986} to compute a spherical correlation coefficient
\begin{equation}
    \hat{\rho}_{X Y}=\frac{\det\left(\frac{\sum_{i}X_{i}Y_{i}^{T}}{n}\right)}{\sqrt{\det\left(\frac{\sum_{i}X_{i}X_{i}^{T}}{n}\right)\det\left(\frac{\sum_{i}Y_{i}Y_{i}^{T}}{n}\right)}}
\end{equation}
for the $n$ samples $X_i,Y_i$ on the 3-sphere $\mathcal{S}^3$, on which all quaternions reside. We visualize the spherical correlation coefficients for all dynamic joints of the AMASS database (excluding toe bases and the root orientation) in \cref{fig:cc_joints}. The illustration depicted in \cref{fig:cc_kin} provides a better intuition of these correlation coefficients, comparing the correlation coefficients along the kinematic tree and for all joints. Notably, high correlations are observed particularly for joints at different leaves of the kinematic tree, such as for the left and right arm joints. We note that the used correlation coefficient only captures certain dependencies on the 3-sphere.

\subsection{Distribution Gaps for the AMASS Datasets}
\begin{figure}
    \centering
    \includegraphics[width=\columnwidth]{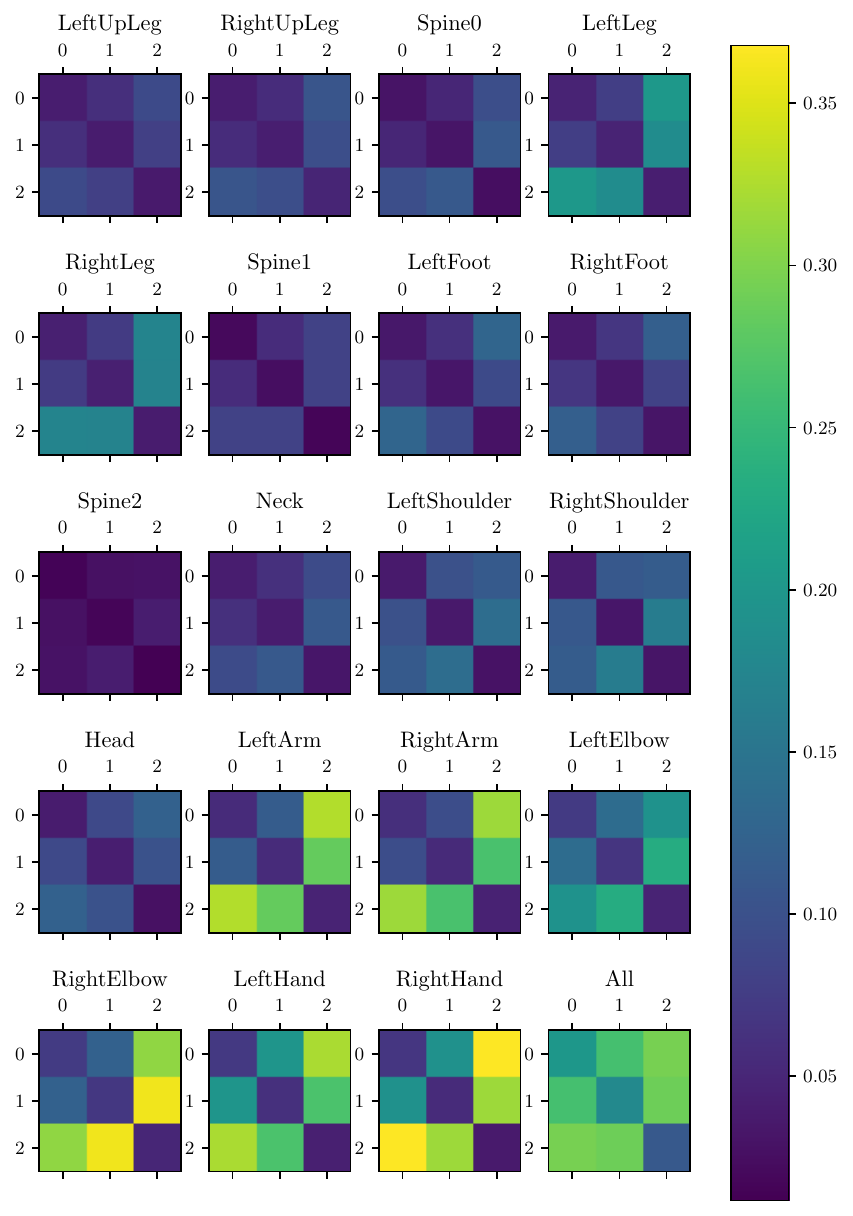}
    \caption{Earth mover's distance between AMASS training (0), validation (1), and test (2) datasets based on the geodesic distance for all dynamic SMPL joints and when comparing \textit{All} joints. For the \textit{All} category, the EMD is computed using the average geodesic distance across multiple joint rotations.}
    \label{fig:emd_joints}
    \vspace{-10pt}
\end{figure}

\begin{figure}
    \centering
    \includegraphics[width=\columnwidth]{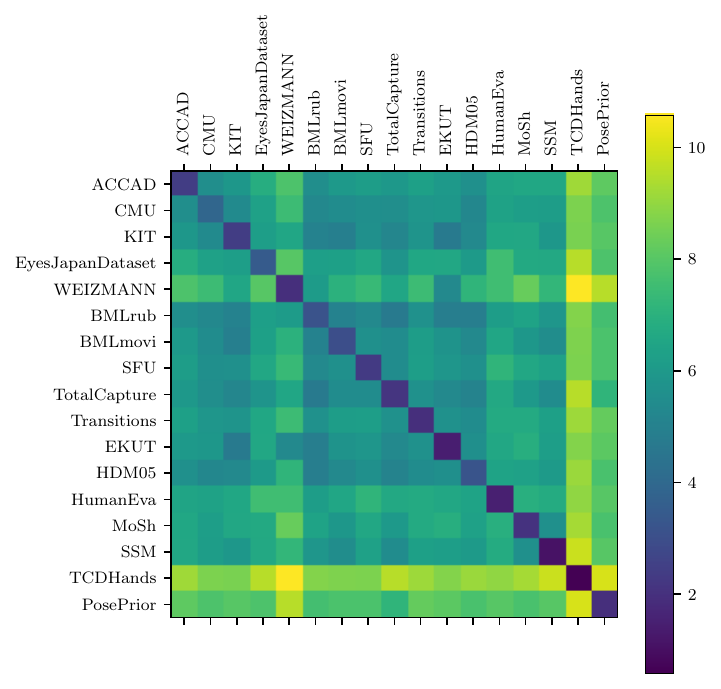}
    \caption{Earth mover distances between various datasets in the AMASS database, computed based on the sum of the geodesic distances of all dynamic joints of the AMASS database.}
    \label{fig:emd_datasets}
    \vspace{-10pt}
\end{figure}

Evaluating an unconditional prior typically assumes \textit{iid} samples in training and test datasets, which is not the case for the AMASS \cite{mahmoodAMASSArchiveMotion2019} database. To compare the distributions of two datasets based on a sample-based similarity metric of the rotations, we compute the earth mover's distance (EMD) \cite{rubner_earth_2000} using the geodesic distance as the distance measure. We compare the EMDs for individual joints and for all joints for the common AMASS datasets split \cite{mahmoodAMASSArchiveMotion2019} in \cref{fig:emd_joints} and the EMD between various datasets of the AMASS database in \cref{fig:emd_datasets}. For computational reasons, we use only $2$k samples for the transport problem, acknowledging potential inaccuracies in higher dimensions. Nevertheless, the distribution gap is highlighted because the entries of the correlation matrix that correspond to the auto-correlation are markedly lower than the cross-correlation.
This aligns with the notable performance disparity between training and test datasets shown in the main paper (\cref{tab:combined_pr_results}).

\section{Implementation and Training Details}
In the following, we elucidate the implementation and training details for HuProSO3 and the integration of the other evaluated methods. 

\subsection{HuProSO3}
We trained the following HuProSO3 models: one as an  unconditional prior, one for inverse kinematics, one for inverse kinematics with randomly occluded 3D key points, and one for 2D to 3D uplifting.

\paragraph*{Architecture.}
The normalizing flow architecture for learning the density $p(\mathbf{R})$ is consistent across all experiments.
We use 12 Möbius coupling layers, with each layer (except the final one) followed by a quaternion affine transformation, totaling 11 quaternion affine layers.  
The parameters of the Möbius transformation are computed by an MLP $g_\text{c-M}(\cdot)$ with three hidden layers and ReLU activations and a hidden dimension of 16.
For the presented applications that require conditioning (inverse kinematics and 2D to 3D uplifting), an MLP $\mathbf{c}=g(\mathbf{c}_\text{feat})$ computes the relevant features from the input context vector $\mathbf{c}_\text{feat}$. We use an MLP with one hidden layer and a hidden dimension of 64. The output dimension of the feature is $\mathbf{c}\in\mathbb{R}^{64}$. 
The complete model that is conditioned on the 3D pose has around 1.5 million parameters.

\paragraph*{Training.}
Our models are trained with a batch size of $1$k, utilizing the Adam optimizer. We set the initial learning rate to 5e-3 and employ a step learning rate scheduler with a multiplicative factor of 0.5 for learning rate decay. 

\paragraph*{Run time.}
Sampling from an autoregressive model is slow and scales with the number of dimensions. 
Parallelizing the evaluation of the model is in general possible. However, our current implementation does not support this. Therefore, evaluating one batch requires around 1.1s on a NVIDIA A40 GPU, while sampling one batch takes around 2.1s.

\subsection{Learning and Optimizing Baseline Methods}

We compare our method with implementations of VPoser~\cite{pavlakos_expressive_2019}, GAN-S~\cite{davydov_adversarial_2022}, Pose-NDF~\cite{avidan_pose-ndf_2022}, and HuManiFlow~\cite{sengupta_humaniflow_2023}. We use the pre-trained models for the priors VPoser, GAN-S, and Pose-NDF. The normalizing flow based on the 6D representation is implemented using the \textit{CircularAutoregressiveRationalQuadraticSpline} module of the normflows libary \cite{stimper_normflows_2023}, with the PDF defined on $\mathbf{x}\in\mathbb{R}^{6\cdot 19}$. 

For conditional tasks, we optimize GAN-S and Pose-NDF for inverse kinematics and 2D to 3D uplifting as outlined in the main paper in \cref{sec:exp_ik} based on 21 SMPL joints. For GAN-S, we utilize the L-BFGS optimizer, set the learning rate to 1 and perform 500 iterations. We use the existing Pose-NDF repository to optimize for occluded joints, and we mask the joint positions in case of occlusions as outlined in \cref{eq:loss_masked_ik} in the main paper.

We train the extracted model of HuManiFlow \cite{sengupta_humaniflow_2023} using the Adam optimizer with a learning rate of 5e-5 with a step learning rate scheduler and step factor of 0.5, and a batch size of 500 until convergence. We apply the same masking strategy as for HuProSO3.

\subsection{Visualization Techniques}
In \cref{fig:huproso_2d}, we adopt the visualization technique introduced by \cite{murphy_implicit-pdf_2021} to display samples on SO(3). A sample on SO(3) is visualized by projecting it onto a 2-sphere and visualizing the third rotation angle through color coding.

\begin{figure}
    \centering
    \includegraphics[width=\columnwidth]{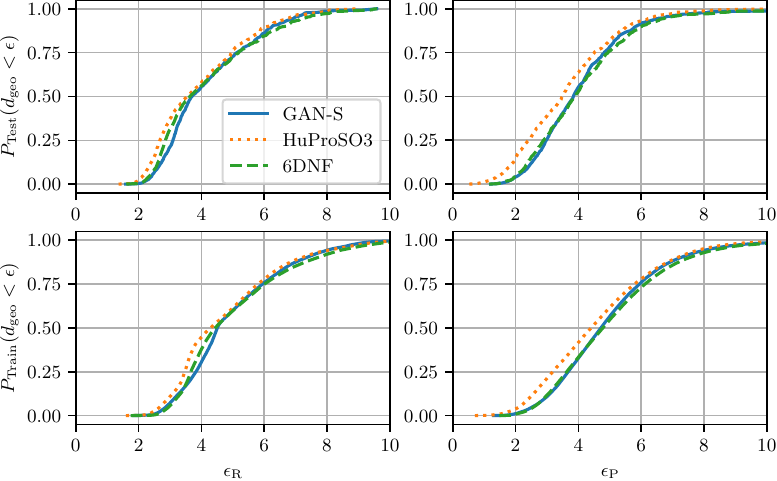}
    \caption{Precision (P) and recall (R) curves for AMASS training and test dataset based on the summed geodesic distances as presented in the main paper, computed for GAN-S, HuProSO3, and 6D normalizing flow. Higher values indicate better quality in all charts.}
    \label{fig:ex_recall}
\end{figure}
\begin{figure}
    \centering
    \includegraphics[width=\columnwidth]{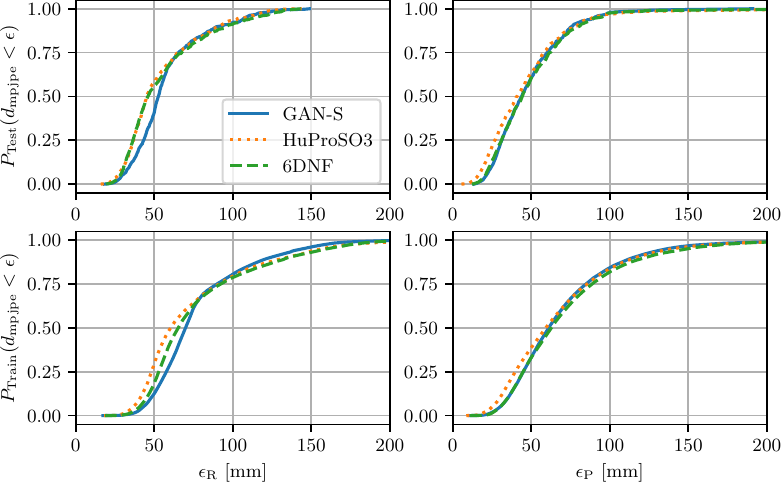}
    \caption{Precision (P) and recall (R) curves for the AMASS training and test datasets based on the MPJPE metric, computed for GAN-S, HuProSO3, and 6D normalizing flow. Higher values indicate better quality in all charts.}
    \label{fig:prec_rec_curve_mpjpe}
    \vspace{-10pt}
\end{figure}

\begin{figure}
    \centering
    \includegraphics[width=\columnwidth]{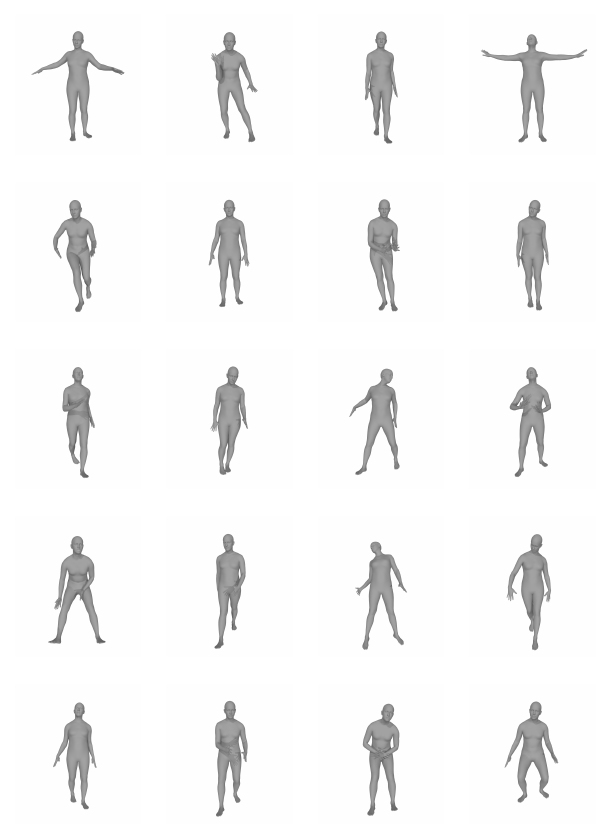}
    \vspace{-10pt}
    \caption{Renderings of randomly sampled human poses from the unconditional HuProSO3 prior.}
    \label{fig:humans_rendered}
\end{figure}

\section{Additional Qualitative Results}
\subsection{Correlation Coefficients for Learned Prior}
To demonstrate that our prior has effectively learned correlations between different joint rotations, we plot the spherical correlation coefficients computed on sampled poses from the prior in figure \cref{fig:cc_joints} next to spherical correlation coefficients computed from samples of the datasets.

\subsection{Precision Recall Curves}
To evaluate the priors, we compute the precision and recall curves. We plot the curves in \cref{fig:ex_recall}.
To assess the priors, we compute and plot the precision and recall curves, as shown in \cref{fig:ex_recall} and \cref{fig:prec_rec_curve_mpjpe}. This comparison, similar to the one presented in \cref{tab:combined_pr_results} of the main paper, is based on the MPJPE metric. Additionally, the mean and median values of these metrics are detailed in \cref{tab:combined_pr_results_mpjpe}.

\begin{table}
\caption{Summary of precision and recall statistics for the AMASS database \cite{mahmoodAMASSArchiveMotion2019}, both on test and training datasets. The values indicate the MPJPEs [mm] for all SMPL joints between samples from the dataset and the evaluated pose prior after applying forward kinematics.}
    \centering
    \footnotesize
    \begin{tabular}{@{}lcc|ccc}
        \toprule
        & \multicolumn{2}{c}{Test (mean [median])} & \multicolumn{2}{c}{Train (mean [median])} \\
        \midrule
        & Recall & Precision & Recall & Precision \\
        \midrule 
        GAN-S \cite{davydov_adversarial_2022} & 50.6 [48.4] & 68.7 [64.2] & 39.7 [35.0] & 56.2 [49.7] \\
        6D NF & 46.3 [\textbf{39.9}] & 65.8 [57.8]  & 39.4 [34.2] & 57.2 [49.4] \\
        Ours & \textbf{45.4} [40.2] & \textbf{61.2 [52.1]}  & \textbf{34.3 [28.8]} & \textbf{51.3 [43.2]} \\
        \bottomrule
    \end{tabular}
    \label{tab:combined_pr_results_mpjpe}
    \vspace{-10pt}
\end{table}

\subsection{Rendered Samples from Unconditional Prior}\label{sec:samples_prior}
To demonstrate HuProSO3's capability as a generative model for sampling realistic and diverse human poses, we present renderings of randomly selected samples in \cref{fig:humans_rendered}.

\subsection{Rendered Samples from 2D to 3D Uplifting} \label{sec:samples_2d_to_3d}
\begin{figure}
    \centering
    \includegraphics[width=\columnwidth]{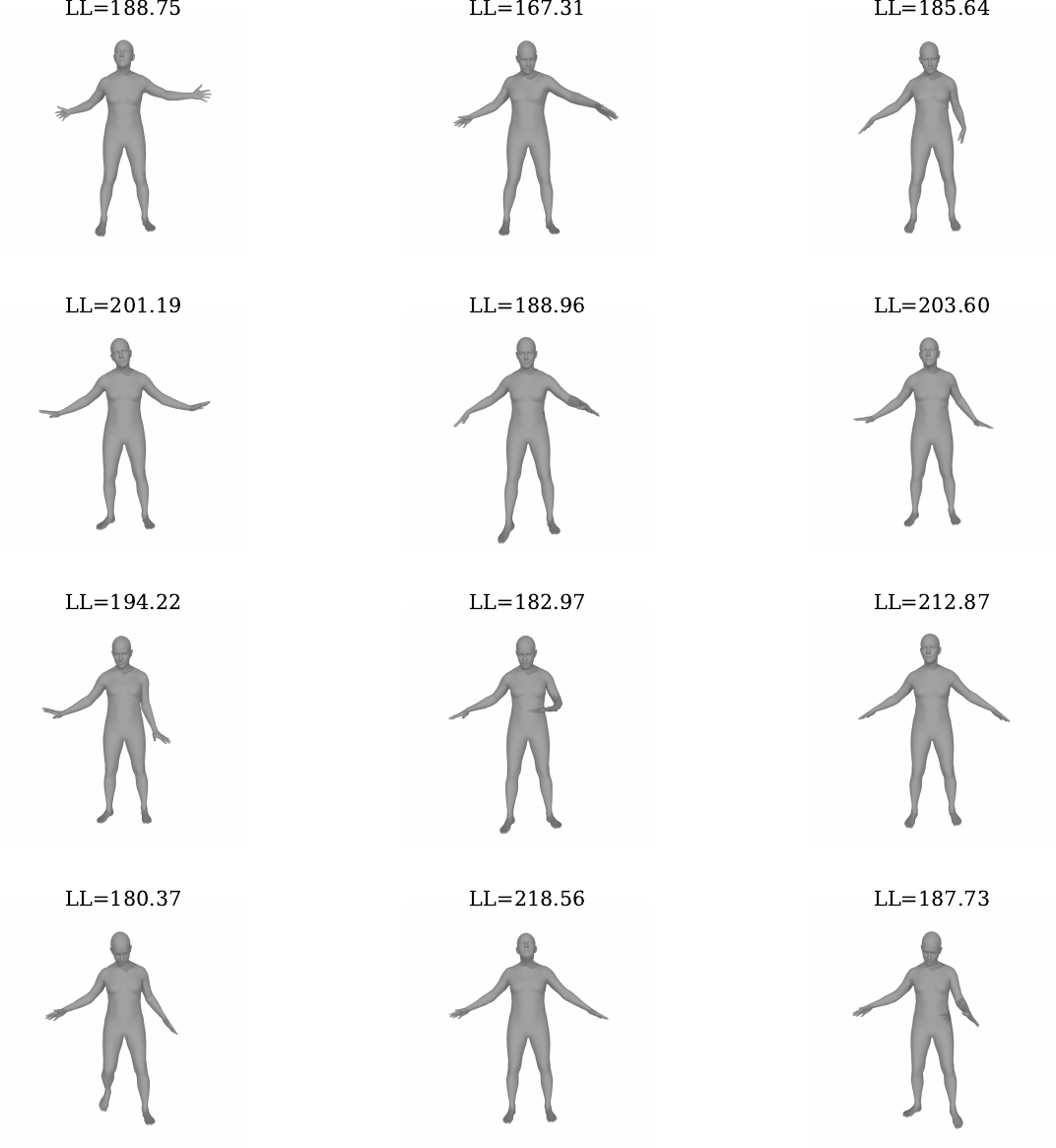}
    \vspace{-10pt}
    \caption{Renderings of human poses sampled from the predicted distribution, conditioned on the 2D key point positions of all joints, excluding the left arm and the right leg. The log likelihood corresponding to a normalized density is displayed above the considered sample.}
    \label{fig:humans_rendered_uplifting}
    \vspace{-5pt}
\end{figure}

Based on the setting and conditioning in \cref{fig:huproso_2d}, we render 12 poses that are sampled from the learned distribution $p(\mathbf{R}|\mathbf{c})$ in \cref{fig:humans_rendered_uplifting}, where $\mathbf{c}$ is derived from 2D key points with the left arm and the right leg occluded. These samples, displayed in \cref{fig:humans_rendered_uplifting}, reveal that while the model often predicts straight right legs, the left arm's pose varies significantly, which follows the training dataset's distribution. For joints where the given 2D key point positions allow inferring their rotations, the estimates show less variability, as we also visualize in \cref{fig:huproso_2d} of the main paper for the standard deviation of the joint's positions and rotations.

Optimization-based methods are limited to inferring a single pose from occluded key points. While this approach might yield accurate results on average, as reflected in mean metrics, it fails to capture the inherent ambiguity in these scenarios.

\section{Additional Quantitative Results}

\begin{table}
\caption{Log likelihood evaluation for inverse kinematics (IK), rotation distribution estimation given 2D key points (2D to SO(3)), and an unconditional prior. Unless specified otherwise, results pertain to the AMASS test dataset.}
    \centering
    \footnotesize
    \begin{tabular}{@{}lccll}
    \toprule
     Method&  IK&2D to SO(3) &Prior&Prior (Train)\\
     \midrule
     HuManiFlow \cite{sengupta_humaniflow_2023}&  100.8& 83.6& -&-\\
     Ours& 217.5&202.2& 137.6&184.7\\
     \bottomrule
    \end{tabular}
    \label{tab:ll_evals}
\end{table}

\subsection{Evaluation of Likelihood}\label{sec:eval_ll}

We evaluate the likelihoods for the unconditional prior with HuProSO3 and for conditional distributions for HuProSO3 and the HuManiFlow approach in \cref{tab:ll_evals}. For the unconditional prior, the likelihood evaluations also reveal a significant gap between the training and test distributions.

\begin{figure}[t] 
	\centering
	\includegraphics[width=0.8\columnwidth]{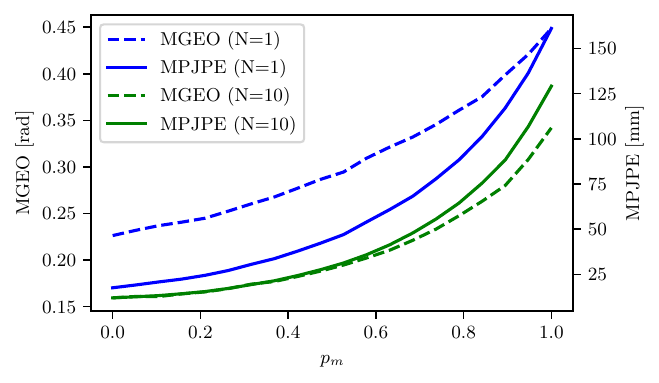}
	\vspace{-5pt}
 \caption{MGEO and MPJPE for a variation of the mask probability $p_\text{m}$ based on one sample and the mean over 10 samples drawn from HuProSO3 conditioned on partially given 3D key point information.}
	\label{fig:vary_pmask}
  \vspace{-5pt}
\end{figure}

\begin{figure}
    \centering
      \begin{subfigure}{0.45\columnwidth}
        \includegraphics[width=0.9\linewidth]{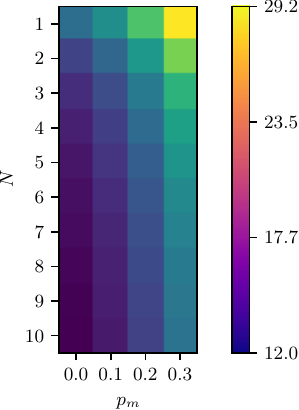}
        \caption{Positional error MPJPE [mm].}
        \label{fig:eval_pm_mpjpe}
      \end{subfigure}
      \begin{subfigure}{0.45\columnwidth}
        \includegraphics[width=0.9\linewidth]{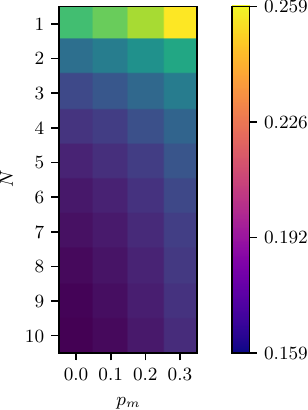}
        \caption{Rotational error MGEO [rad].}
        \label{fig:eval_pm_geo}
      \end{subfigure}
    \caption{MPJPEs and MGEOs for different masking probabilities $p_m$ and numbers of samples from the learned distribution $N$. The metrics are calculated based on the mean of the sampled poses by computing the average rotation for each joint.}
    \label{fig:eval_pm_k}
    \vspace{-5pt}
\end{figure}

\begin{table}
\caption{Summary of precision and recall statistics for the AMASS database \cite{mahmoodAMASSArchiveMotion2019}, both on test and training datasets. The reported values represent the cumulative geodesic distances for all joint rotations between samples from the dataset and the evaluated pose prior. Pose-NDF~1 was optimized using random intialization, Pose-NDF~2 using slightly noised poses from the AMASS test dataset.}
    \centering
    \footnotesize
    \begin{tabular}{@{}lcc|cc}
        \toprule
        & \multicolumn{2}{c}{Test (mean [median])} & \multicolumn{2}{c}{Train (mean [median])} \\
        \midrule
        & Recall & Precision & Recall & Precision \\
        P-NDF 1 & 17.7 [17.7] & 14.5 [14.4]  & 17.5 [17.2] & 14.7 [14.4] \\
        P-NDF 2 & 4.83 [4.77] & 5.63 [5.42]  & 6.11 [5.92] & 6.88 [6.65] \\
        Ours & 3.44 [2.95] & 4.24 [3.71]  & 2.93 [2.64] & 3.90 [3.59] \\
        \bottomrule
    \end{tabular}
    \label{tab:pr_results_posendf}
    \vspace{-5pt}
\end{table}

\subsection{Comparison to Pose-NDF as Pose Prior}
We evaluate Pose-NDF as a pose prior by computing the precision and recall statistics. In a first experiment, we evaluate when initializing with noise. However, this does not result in realistic poses since Pose-NDF generates realistic poses when the initialized poses are close to the training distribution. In a second experiment, we add a small amount of noise to poses of the test distribution ($\sigma_\text{noise}=0.1$), which provides realistic poses. However, such an initialization inherently biases the optimization towards in-distribution samples. Therefore, it is highly depending on the similarity between training and test distribution.

\begin{table}
\caption{Summary of precision and recall statistics for the AMASS database \cite{mahmoodAMASSArchiveMotion2019}, both on test and training datasets. The reported values represent the cumulative geodesic distances for all joint rotations between samples from the dataset and the evaluated pose prior.}
    \centering
    \footnotesize
    \begin{tabular}{@{}lcc|ccc}
        \toprule
        & \multicolumn{2}{c}{Test (mean [median])} & \multicolumn{2}{c}{Train (mean [median])} \\
        \midrule
        & Recall & Precision & Recall & Precision \\
        \midrule 
        GAN-S \cite{davydov_adversarial_2022} & 3.76 [3.34] & 4.51 [4.23] & 3.57 [3.34] & 4.38 [4.13] \\
        6D NF & 3.66 [3.16] & 4.50 [4.00]  & 3.55 [3.32] & 4.42 [4.10] \\
        Ours & 3.44 [2.95] & 4.24 [3.71]  & 2.93 [2.64] & 3.90 [3.59] \\
        \bottomrule
    \end{tabular}
    \label{tab:asdasd}
    \vspace{-5pt}
\end{table}

\begin{table}
\caption{Comparison to GFPose-rot: Minimum MGEO and minimum MPJPE are computed based on 20 generated samples for the occlusion of leg (L), arm and hand  (A), and upper arm (S). The results are presented for $10$k random samples from the AMASS test datasets. The GEO metrics are averaged over 21 joints.}
    \centering
    \footnotesize
    \begin{tabular}{@{}l|cccccc}
    \toprule
     Method & \multicolumn{3}{c}{minMGEO} & \multicolumn{3}{c}{minMPJPE} \\
     \midrule
     Occlusion &  L & A & S &  L & A & S\\
     \midrule
     GFPose-rot (N=1)&  0.104 & 0.108& 0.089& 7.9& 14.3& 4.9\\
    GFPose-rot (N=20)& 0.103& 0.107& 0.088&7.8& 14.1& 4.8\\
    Ours (N=1)& 0.217&0.254&0.208&20.8& 39.0& 18.7\\
    Ours (N=20)& 0.070& 0.081& 0.067&5.7& 11.2& 5.2\\
    \bottomrule
    \end{tabular}
    \label{tab:comparison_to_GFPoseRot}
\vspace{-10pt}
\end{table}

\subsection{Comparison to GFPose-rot}

Directly comparing to GFPose \cite{ci_gfpose_2022} is not possible since it was not trained on the AMASS database and it is parameterized with joint positions. Therefore, we adapt the implementation of GFPose and we train it on the AMASS database using the axis-angle representation using the same hyperparameters as in the original repository.
For the occlusions, we apply the same masking strategy as in our implementations. We follow \cite{ci_gfpose_2022}  and evaluate using the minimum error sample. We use 10 times fewer samples than GFPose (N=20) and we report the results for minimum geodesic distance and joint position error in \cref{tab:comparison_to_GFPoseRot}. 
In our experiments, GFPose-rot collapses to the mean pose. While our results are worse for single sample evaluations, our model provides more diverse samples than GFPose-rot.

Here, a disadvantage of our model becomes apparent: Since our base distribution is uniform on SO(3), computing the mode as when considering a Gaussian distribution is not possible. This might be a reason, why generating an individual sample does not achieve competing results.

While GFPose-rot achieves partially better results, it does not provide a normalized density.

\subsection{Per-Vertex Errors for Inverse Kinematics and Occluded Joints} \label{subsec:per_vertex_error_ik}
We compare HuProSO3's per-vertex error performance with HuMoR \cite{rempe_humor_2021} using \textit{TestOpt}, VPoser \cite{pavlakos_expressive_2019}, and Pose-NDF \cite{avidan_pose-ndf_2022} for 60 frames, following the protocol in \cite{avidan_pose-ndf_2022}. 
We present the per-vertex error results for inverse kinematics and occluded joints in \cref{tab:pve_occ}. While the optimization-based methods achieve a better performance for the MPJPE metric, the presented results in \cref{tab:pve_occ} also support that the wrong rotation estimates result in worse performance when comparing all mesh vertices on the rendered human.
\begin{table}
 \caption{Comparison of per-vertex errors [mm] across various occluded joints in the AMASS test dataset: left leg (L), left arm and hand (A+H), and right shoulder and upper arm (S+UA). The results are based on 60 frames as reported in \cite{avidan_pose-ndf_2022}.}
    \centering
    \footnotesize
    \begin{tabular}{@{}lccc}
        \toprule
         Method&  L&  A+H&S + UA\\
         \midrule
         VPoser \cite{pavlakos_expressive_2019}&  25.3&  85.1&99.8\\
         HuMoR \cite{rempe_humor_2021}&  56.0&  78.3&47.5\\
         Pose-NDF \cite{avidan_pose-ndf_2022}&  \textbf{24.9}&  78.1&76.3\\
         \midrule
         Ours (N=10)& 34.0 & \textbf{57.5} & \textbf{34.5}\\
         \bottomrule
    \end{tabular}
    \label{tab:pve_occ}
\vspace{-10pt}
\end{table}

\subsection{Evaluations for Varying Numbers of Samples}\label{subsec:vary_number_samples} 
We present additional evaluation results for varying numbers of samples that are used for computing positional and rotational errors in \cref{fig:eval_pm_k} and \cref{fig:vary_pmask}.  \cref{fig:eval_pm_k} extends the visualizations from \cref{fig:vary_pmask} by presenting MPJPE and MGEO metrics for different masking probabilities and sample counts drawn from HuProSO3, with each joint masked with a probability of $p_m$. 

\textbf{Baselines and Setup.}
In addition to HuProSO3, as detailed in the main paper, and the model based on the HuManiFlow implementation, we also compare with an implementation that uses normalizing flows defined on SO(3) as for HuProSO3, but with fixed ancestor-conditioning as suggested in \cite{sengupta_humaniflow_2023}. We train this model following the same strategy as for HuProSO3.  For evaluation, joints are randomly masked with a probability of $p_m=0.3$. We present the errors for the average joint rotations (as in the main paper) and, additionally, based on the pose out of the $N$ sampled poses that results in the lowest considered error metric (minimum error sample). We compute all results based on $10$k randomly drawn samples from the AMASS test dataset.

\textbf{Discussion.}
Our analysis reveals that normalizing flows designed with the SO(3) modeling approach seem to more effectively capture the joint distribution than the approach by \cite{pmlr-v89-falorsi19a} applied in \cite{sengupta_humaniflow_2023}.
While the error for a single sample ($N=1$) is similar across both ancestor-conditioned models (\textit{SO3 AC} and \textit{HF AC}) for the selected masking strategy, the SO(3)-based model yields lower minimum errors with an increased number of samples.
However, ancestor-conditioning along the kinematic tree does not fully capture all statistical dependencies. Consequently, HuProSO3, which is not limited to fixed conditioning sequences and operates on the product space of all SO(3) manifolds, demonstrates notably lower minimum and average pose errors, reflecting its superior capability in learning dependencies of the joint rotations.

\end{document}